\documentclass[10pt,journal,compsoc]{IEEEtran}
\usepackage{cite}
\usepackage{amsmath,amssymb,amsfonts}
\usepackage{algorithmic}
\usepackage{graphicx}
\usepackage{textcomp}
\usepackage{xcolor}
\usepackage{amsmath}
\usepackage{mathtools}

\DeclarePairedDelimiter\floor{\lfloor}{\rfloor}

\usepackage{algorithmic}
\usepackage{graphicx}
\usepackage{textcomp}
\usepackage{xcolor}
\usepackage{float}
\usepackage{wrapfig}
\usepackage{array}
\usepackage{multirow}
\usepackage{hyperref}
\usepackage{comment}
\usepackage{enumitem}  
\hypersetup{
    colorlinks=true,
    linkcolor=blue,
    filecolor=magenta,      
    urlcolor=blue,
}

\newcolumntype{L}[1]{>{\raggedright\let\newline\\\arraybackslash\hspace{0pt}}m{#1}}
\newcolumntype{C}[1]{>{\centering\let\newline\\\arraybackslash\hspace{0pt}}m{#1}}
\newcolumntype{R}[1]{>{\centering\let\newline\\\arraybackslash\hspace{0pt}}m{#1}}

\ifCLASSINFOpdf
\else
\fi
\begin{document}

\title{Identification of Stochasticity by Matrix-decomposition: Applied on Black Hole Data}
%
%



\author{
  \IEEEauthorblockN{Chakka Sai Pradeep, Sunil Kumar Vengalil, Neelam Sinha\\}
  \IEEEauthorblockA{International Institute of Information Technology, Bangalore, India\\}

\IEEEcompsocitemizethanks{\IEEEcompsocthanksitem E-mail: saipradeep.chakka@gmail.com, vengalilsunilkumar@gmail.com, neel.iam@gmail.com}}

\IEEEtitleabstractindextext{%
\begin{abstract}

Timeseries classification as stochastic (noise-like) or non-stochastic (structured), helps understand the underlying dynamics, in several domains. Here we propose a two-legged matrix decomposition-based algorithm utilizing two complementary techniques for classification. In Singular Value Decomposition (SVD) based analysis leg, we perform topological analysis (Betti numbers) on singular vectors containing temporal information, leading to SVD-label. Parallely, temporal-ordering agnostic Principal Component Analysis (PCA) is performed, and the proposed PCA-derived features are computed. These features, extracted from synthetic timeseries of the two labels, are observed to map the timeseries to a linearly separable feature space. Support Vector Machine (SVM) is used to produce PCA-label. The proposed methods have been applied to synthetic data, comprising 41 realisations of white-noise, pink-noise (stochastic), Logistic-map at growth-rate 4 and Lorentz-system (non-stochastic), as proof-of-concept. Proposed algorithm is applied on astronomical data: 12 temporal-classes of timeseries of black hole \textit{GRS 1915 + 105}, obtained from RXTE satellite with average length 25000. For a given timeseries, if SVD-label and PCA-label concur, then the label is retained; else deemed ``Uncertain". Comparison of obtained results with those in literature are presented. It’s found that out of 12 temporal classes of \textit{GRS 1915 + 105}, concurrence between SVD-label and PCA-label is obtained on 11 of them.

\end{abstract}

}

\maketitle

\IEEEdisplaynontitleabstractindextext

\IEEEpeerreviewmaketitle

\IEEEraisesectionheading{
\section{Introduction}\label{sec:introduction}}

In several domains, such as weather, finance, agriculture, astronomy etc., studies are performed by collecting measurements and creating models aimed at explaining the underlying phenomena that resulted in the data. Several of these measurements are captured as timeseries, which is a time-ordered sequence of values. The first step towards examining the timeseries is to determine if the timeseries is stochastic, which would mean that it is noise; or if the timeseries is non-stochastic, in which case, physical attributes explaining them would be critical. As is well-studied in a stochastic timeseries, any realization is independent of all other realizations. Standard examples of well-studied stochastic timeseries are white noise, pink noise, etc. On the other hand, standard examples of non-stochastic timeseries are logistic map (at growth rate = 4), Lorentz system. They result in timeseries that exhibit a well-defined structure. For a non-stochastic timeseries, there would be associated physical phenomena, a set of factors that interplay bringing about a well-defined structure. This classification is important in scenarios where the possible influences are largely unknowable and inaccessible, such as astronomy.

It is well-acknowledged that the study of black hole sources is a challenging problem in astronomy. In order to identify a black hole, one needs to study its environment. The immediate neighborhood is often a disc-like structure that is formed due to the in-falling matter, called accretion disc. The timeseries of black hole source, measured at different times (different temporal classes), could exhibit drastically varying behavior (temporal, spectral), requiring different models to explain them. It is important to identify which of the temporal classes are stochastic and which are non-stochastic, to further the understanding of the black hole source. 

Here, we explore the black hole source \textit{GRS 1915+105}. In \cite{newgrsRef}, authors have described the obtained data as exhibiting drastic intensity changes over various timescales, ranging from seconds to days. Of interest, was the observation that the source exhibited quasi-periodic brightness sputters with varied duration over diverse timescales. Such occurrences were said to be occasional, while more common variations were seen to be the faster ones, with smaller amplitudes. These observations make the analysis challenging. This source has 12 different temporal classes: $\alpha$, $\beta$, $\gamma$, $\delta$, $\lambda$, $\kappa$, $\mu$, $\nu$, $\rho$, $\phi$, $\chi$ and $\theta$ \cite{Belloni}.

The temporal classes represent the diverse timeseries measured from the same source that exhibits varied characteristics at different times. Variations across these classes lie in level of X-ray emission, particularly in different energy bands. Also they exhibit different combinations of quasi-periodic oscillation (QPO) frequencies and power spectral density (PSD) structure. The study in \cite{Belloni} explained the differences across the temporal classes and their implications. A brief description of each of these temporal classes is outlined below.
\begin{enumerate}
	\item Class $\alpha$: A low-frequency QPO and a high-frequency QPO, as well as a steady, high level of X-ray emission.
	\item Class $\beta$: A steady, high level of X-ray emission and a low-frequency QPO.
	\item Class $\gamma$: A steady, high level of X-ray emission and a high-frequency QPO.
	\item Class $\delta$: A steady, high level of X-ray emission, a low-frequency QPO, and a high-frequency QPO.
	\item Class $\lambda$: A series of rapid X-ray flares separated by periods of low emission, and a low-frequency QPO.
	\item Class $\kappa$: A series of rapid X-ray flares separated by periods of low emission, and a high-frequency QPO.
	\item Class $\mu$: A series of rapid X-ray flares separated by periods of low emission, a low-frequency QPO, and a high-frequency QPO.
	\item Class $\nu$: Moderate level of X-ray emission, a stable light curve, and a low-frequency QPO.
	\item Class $\rho$: Moderate level of X-ray emission, a stable light curve, and a high-frequency QPO.
	\item Class $\phi$: Moderate level of X-ray emission, a stable light curve, a low-frequency QPO, and a high-frequency QPO.
	\item Class $\chi$: Very low level of X-ray emission, a stable light curve, and a low-frequency QPO.
	\item Class $\theta$: Very low level of X-ray emission, a stable light curve, and a high-frequency QPO.
	
\end{enumerate}
The objective of the present study is to understand the dynamics of the system giving rise to these measurements. Towards this, the first step is to determine if the label that each of these temporal classes belongs to, is stochastic or non-stochastic. The implication of a temporal class belonging to stochastic label would be that the underlying dynamics is noise or that noise has corrupted the true signal to such an extent that it is not distinguishable from noise. On the other hand, if the temporal class is determined as having the label non-stochastic, the underlying dynamics could be attributed to some specific phenomenon that would need to be understood using system identification parameters.

\begin{figure*}[h]
  \includegraphics[width=\linewidth]{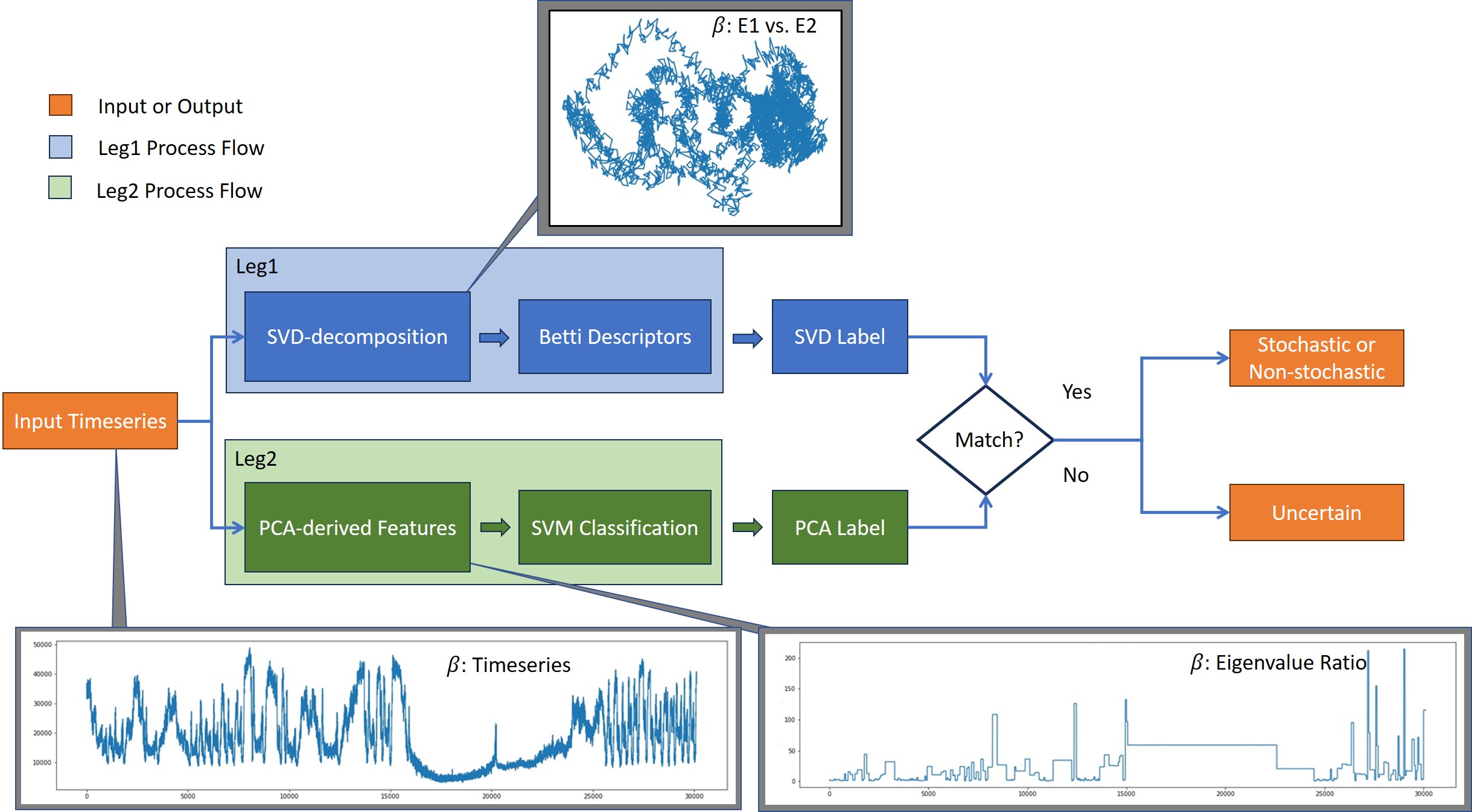}
  \caption{System Overview Illustrated on \textit{GRS 1915 + 105} source $\beta$ timeseries.}
  \label{overview}
\end{figure*}

\section{Related Work}

The task of timeseries classification could involve varying aspects based on the application. Generically, the task of classification requires one to abstract characteristics specific to each classification label. A relatively simpler scenario occurs when standard templates are available for each of the labels, or when they can be assumed to lie within linear sub-spaces. In such cases, similarity-based approaches, clustering techniques are popular. In certain other scenarios, the raw data might need to be transformed to another domain in order to enable analysis. Transforms such as Fourier, Wavelets, Shapelets are popular choices.

Timeseries are often dealt with in audio processing, which includes tasks such as speaker identification, language recognition, speech retrieval, music processing, etc. \cite{dtw_speech, wordrecog}. These tasks are generally performed using frequency-domain analysis. Movements of living beings are recorded as trajectories in 3D space leading to vector-valued timeseries. Examples include insect wing beating, human gait processing, sign language understanding, gesture recognition etc. Here, it is common to utilize signal processing notions of shapelets and multi-resolution analysis. For applications such as online hand-writing identification, signature identification \cite{writing}, the most common approach is to use shape descriptors. For anamoly detection in biomedical signals such as ECG \cite{ecg_ts}, EEG \cite{eeg_tetc1, eeg_tetc2}, fMRI timeseries analysis \cite{eeg_ts}, transform-domains such as wavelets and notions of deconvolution, ICA, etc. are popular. In several of these instances, the objective is to compare a given test instance against a set of standard templates, in order to determine the classification label. The distance measure to determine the classification label could be on the raw timeseries or on the feature vectors derived from the timeseries \cite{ecg_morph}. Similarity measure based classification techniques include popular approaches such as DTW \cite{origDTW_68} and its several variants. 
  
Another popular approach is to view the timeseries using a transformed representation. An example of this transformation is a Hopfield neural network which is used for information retrieval. Hopfield neural network \cite{hopfield} based approaches utilize the notion that each distinct class is represented by a specific configuration of the neurons and weights. These configurations are mapped to energy wells, which are then exploited for classification. Several of these techniques are readily available in packages \cite{pyts} that can be easily installed and utilized by the user. Yet another alternate representation is to view the timeseries as a graph. Graph-based features can be exploited to gain insights and perform classification. Visibility algorithms have built connections between timeseries analysis, graph theory and non-linear dynamics. The authors of \cite{lacasa2010} proposed horizontal visibility algorithm to classify various timeseries. One of the common applications of PCA, which is to identify features that are dominant, has also been used to refine the classification process. In \cite{Silva2022}, authors utilized PCA to determine which of the topological properties of the graph need to considered in the subsequent process of clustering. Hence an unsupervised approach to obtain the desired labeling was reported on diverse classes of natural and man-made signals available in open repositories.



However, a difficult classification scenario occurs when the characteristics that distinguish one label from another are not easy to abstract, with no available standard templates, and limited access to data, as in the task of stochastic vs. non-stochastic classification.

\textbf{Stochastic vs. Non-stochastic classification:} A stochastic timeseries is defined as one where each of the sample is independent of all other samples. Such timeseries are attributed to noise, with no structure within them. On the other hand, a non-stochastic timeseries is characterized by the presence of some well-defined structure. There are no available standard templates to compare a given timeseries to determine which of the labels it belongs to. Hence, it is required to devise entirely different approach for this classification task. It is required to quantify notions of independence across various timestamps and similarity (or lack of) across various timescales. In reported works in literature as summarized in survey reports such as \cite{datadriven}, it is seen that diverse notions such as phase transitions, Correlation Dimension (CD), Entropy-based approaches, Compressed Sensing etc have been utilized for this task. The most popular method to classify a timeseries as stochastic or non-stochastic is Correlation Integral (CI)-based. This approach involves the determination of CD for varying values of Embedding Dimension ($M$), as outlined in \cite{CIGRacia}. In order to analyze a timeseries using the CI approach, one needs to compute CD for specific ranges of $M$. Estimates of the required time-lag $\tau$ are generally obtained using the Autocorrelation function. Typically the value where Autocorrelation function reaches zero or where the first local minimum is encountered, is taken as the value of $\tau$. The value of $M$ is an estimate of how many independent factors contribute to the underlying dynamics. The value of $\tau$ is the number of time-stamps needs to be skipped (time-lag) in order to make the measurements independent. This results in a data matrix, where each column is an observation vector with $M$ components, such that each of the components are independent. Each row contains $k$ consecutive samples taken from the time series. Correlation sum is then computed for various values of ``radius" and embedding dimension, $M$, as explained in \cite{CIGRacia}. This method is known to be computationally expensive, $O(N^2M)$, where $N$ is the length of the timeseries. Several variants \cite{chris_1915, corana} of this approach have emerged over the years, making the computation efficient and robust.

Entropy of random variable, known to be a measure of disorder, is an indicator of the information in the random variable. Hence, entropy, its variants such as approximate entropy, mutual information measures are popular tools to accomplish this classification task. Works reported in \cite{splrecent, russian, Boaretto2021} exploit the idea of an approximate entropy through recurrence plots and phase-space reconstruction. It is well known that such reconstructions require thorough domain knowledge or several empirical trials to estimate the best values of the parameters involved. They are computationally expensive since additional parameters such as Lyapunov exponent need to be computed. One variant in utilizing entropy is the idea of ordinal subsequences which allow computation of Permutation Entropy (PE). This idea was reported in determining complexity measure of timeseries in \cite{Bandt2002}. This work was further extended by the inclusion of a neural network based classifier in the work reported in \cite{Boaretto2021}. Well known synthetic signals were parameterized to learn the distribution of the devised parameter, in an attempt to create templates for the two labels of stochastic and non-stochastic timeseries. A test timeseries undergoes the same parameterization process, and its deviation from the set templates is measured to arrive at its label. The performance of this methodology was illustrated on synthetic signals such as various types of noise and empirical datasets containing human heart rate variability, gait data etc. The only assumption in this approach is prior knowledge on the length of ordinal sequences. A more recent work explored the idea of PE coupled with recurrence of microstates \cite{Prado_2022} to determine if data is stochastic or not. This was applied on diverse data such as sound vocalization and human speeches, heart beat data, electronic circuit, neuronal activity, and geomagnetic indexes.

Another set of approaches exploits compressive sensing, utilizing the basic principle that the dynamics of several  systems (both, natural and man-made) are governed by a small set of functions, that can be estimated using simple expansions. The problem of determining the label is posed as coefficient estimation where the computations are optimized to yield sparse vectors. An appropriate choice of basis becomes critical for this set of approaches to be effective. The study reported in \cite{Brunton2016} determined the dynamics by exploring ideas of sparsity and machine learning in non-linear dynamical or dynamic systems. In this work, the authors exploit the empirical observation that there are very few functions in the available function space that govern the dynamics of most physical phenomena. Hence choice of appropriate basis should result in sparse representation. The performance is illustrated on well-studied systems such as Lorentz systems, and is shown to be robust to noise. 

The application of Machine Learning techniques for this classification task is now being extensively reported. The study in \cite{hivecote} reports the utility of ensemble models in timeseries classification on data from publicly available archives. The methodology utilizes signal processing concepts such as shapelets as well as notions of Bag-of-words to achieve the classification, on data emanating from physical phenomena as varied as agriculture, medicine and wild-life, where the typical length of these timeseries is of the order of hundreds. Recent success of deep learning in computer vision tasks has made it an exploratory tool for several aspects of data analysis including the problem at hand. Deep learning for timeseries classification, as summarized in the survey paper \cite{review2019paper}, is being explored by several research groups. Generative models such as autoencoders are used, along with CNNs and LSTMs. Discriminative models based on architectures such as ResNet, variations of LeNet, etc are also explored mainly for efficacy in feature engineering. The utility of autoencoders to represent the timeseries in latent space has been explored. The work in \cite{ds_icassp} approaches this problem by quantifying the extent of stochasticity in a timeseries, by utilizing autoencoder-based time-invariant representation. This is accomplished by reconstructing the given timeseries over several time-scales. A stochastic timeseries would not be sensitive to the time-scale while a non-stochastic one would be severely affected by the choice of time-scale. The present authors have applied their approach on the RXTE dataset, in order to determine which of the temporal classes are stochastic and which others are non-stochastic. Yet another work reported in \cite{lss} accomplishes the task of classification by converting a timeseries to a 2D representation called ``Latent space signature" (LSS), which plots the tuples of latent-space representation simultaneously obtained using time- and frequency-domain analyses. The present work reports the performance of their approach on the same RXTE dataset, to accomplish the same task of determining which of the temporal classes are stochastic and which are non-stochastic.

In this work, we propose matrix-decomposition based technique with lower computational complexity, as outlined in the subsequent section.

\section{Proposed Method: Matrix Decomposition based }
	In this work, we propose an algorithm with two parallel legs as shown in Fig. \ref{overview}, each of which performs complementary analysis, in order to determine if the given timeseries is stochastic or non-stochastic. They are: (1) SVD decomposition followed by topological analysis (using Betti number descriptors), leading to SVD-label, (2) PCA-derived features followed by SVM-classification, leading to PCA-label. If the SVD-label and PCA-label concur, the corresponding label is retained; else the result is declared ``Uncertain". 
	\subsection{Contributions} Contributions of the paper are:
	\begin{enumerate}
		\item Matrix decomposition based techniques for classification of a timeseries as stochastic or non-stochastic.
		\item Utility of topological descriptors to capture temporal structure in a timeseries
		\item Utility of PCA to determine stochasticity
		\item Combining temporal ordering-specific with ordering-agnostic approach to arrive at the classification label
	\end{enumerate}

All codes and results are made publicly available in the link \href{https://github.com/sunilvengalil/ts_analysis_pca_eig}{https://github.com/sunilvengalil/ts\_analysis\_pca\_eig}

	The proposed method utilizes temporal-ordering specific analysis through SVD decomposition, and ordering-agnostic analysis through PCA analysis. Each of these approaches is described in separate subsections, subsequently. For the description, we utilize, ($z_1, z_2 \mathellipsis z_N$) to denote a  timeseries of scalar values, of length $N$.  

	\subsection{SVD Decomposition - Topological Analysis}
In this leg, we form uncorrelated observation vectors from the raw timeseries data  by choosing an appropriate embedding dimension $M$, and phase lag $\tau$, as in \cite{misra2006}, using autocorrelation plot. Data matrix, $D$, given in equation (\ref{eqn:dmatrix}), is formed with each row as the time-shifted version of the original timeseries, given by
\begin{equation} \label{eqn:dmatrix}
	D= \begin{bmatrix} z_1  & \mathellipsis & z_k \\
		z_{1+\tau}  & \mathellipsis & z_{k+\tau} \\
		\mathellipsis & \mathellipsis & \mathellipsis\\
		z_{1 + (M-1)\tau} & \mathellipsis & z_{k + (M-1)\tau)}\\
	\end{bmatrix}.
\end{equation}
Here $z_{1}, z_{2}, \mathellipsis, z_{k}$ are $k$ consecutive samples of the timeseries. Each column can be viewed as a different observation vector of the same time-evolving phenomenon. Temporal dynamics is understood by utilizing the right singular vectors of the SVD decomposition of $D$ as given by $D = U \Sigma V^T$. The first two columns of matrix $V^T$, which are the top two right singular vectors, E1 and E2, are considered. The topology of the plot E1 vs. E2, for non-stochastic timeseries, is expected to show a specific pattern (structured trajectory with well-defined voids). On the other hand, E1 vs. E2 plot for stochastic timeseries would show no directional preference and hence would result in a single blob without any voids. This difference in the characteristics of E1 vs. E2 plot is captured using Betti descriptors.

\subsubsection{ Betti descriptors for E1 vs. E2 plot} 

Betti number descriptor \cite{topology} for a $d$-dimensional manifold is a vector of $d$ integers which is represented as $B = (B_0, B_1 \mathellipsis B_{d-1})$. Here $B_{0}$ is the number of blobs (connected components) and $B_k$ represents number of $k$-dimensional voids for $k>0$.  The E1 vs. E2 plots are two-dimensional manifolds, which are described by  $B=(B_{0}, B_{1}$). Here, we utilize $L1$-norm of the $B$ vector, $\|B\|_1$, for classification.  For a stochastic time series, $B_{0}$  is expected to be 1 (single connected component) and $B_1$ is expected to be 0 (no voids). Hence for a stochastic timeseries, $\|B\|_1 = 1$. However, for a non-stochastic timeseries, we hypothesize that $B_{0}$ can be greater than 1 (can have more than 1 connected component) and $B_1$ is always greater than 0 (presence of voids) due to the attractor behavior. Hence for a non-stochastic timeseries $\|B\|_1 > 1$. This simple observation is utilized to classify a given timeseries. 

By the end of this analysis leg, a classification label, here called the SVD label, is output. This analysis regards temporal ordering, incorporated right at the stage of forming the data matrix.

\subsection{PCA-derived Features - SVM Classification}

PCA-decomposition is utilized here to infer whether the given timeseries possesses a dominant orientation or not. This is computed by hierarchically splitting the timeseries into two halves, and computing the covariance matrix of these split observations. The eigenvalues of this $2 \times 2$ covariance matrix, are computed and compared. We define the eigenvalue ratio as the ratio of larger eigenvalue to smaller eigenvalue. One of two signatures is observed: If the data indeed show any dominant direction (as in non-stochastic timeseries), then the larger eigenvalue will be significantly greater than the other, resulting in large value of eigenvalue ratio ($>>1$). On the other hand, if the data do not show any dominant direction (as in stochastic timeseries), then eigenvalues of the covariance matrix will be comparable, leading to small eigenvalue ratios ($\approx1$). This observation is utilized in devising features, as outlined below.

For a timeseries consisting of $N$ entries: $(z_1, z_2 \mathellipsis z_N)$, the following is performed.
\begin{enumerate}
	\item Split the series into two halves: $(z_1, z_2 \mathellipsis z_{\floor*{\frac{N}{2}}})$ and $(z_{\floor*{\frac{N}{2}} + 1}, \mathellipsis z_N)$
	\item Compute covariance matrix, $C$,  by treating the samples in two halves as $\floor*{\frac{N}{2}}$ observations of two-dimensional vectors
	\item Find eigenvalues of $C$ as $\lambda_1$ and $\lambda_2$; the eigenvalue ratio is computed as  $\lambda_1/\lambda_2$ where $\lambda_1 \ > \lambda_2$ (eigenvalues of a covariance matrix are real)
	\item If the eigenvalue ratio for an interval is smaller than a value of threshold, $Th$ (computing optimal value of $Th$ is described below), the interval is further split into two sub-intervals of equal size.
\end{enumerate}
 Subsequently, eigenvalue ratio for each sub-interval is computed. The process is repeated as long as length of the sub-interval is greater than a predefined number of samples (here taken as 100). In each iteration of data split, the algorithm attempts to find structure in the timeseries. Each iterative split represents a shorter time-segment implying that the search for structure is now shifted to a smaller time-scale. Eitherway, search is halted when the number of samples reaches less than 100.
 
For a fixed value of $Th$, the following PCA-derived features are proposed:
\begin{itemize}
	\item  Variance of Eigenvalue Ratio ({VER}): Variance of eigenvalue ratios of covariance matrices across sub-intervals in the entire timeseries is computed. This captures spread in eigenvalue ratios, which is expected to be large for non-stochastic timeseries. 
	\item Area Under the Eigenvalue Ratio curve ({AUER}): This denotes area under the curve of eigenvalue ratio for the entire timeseries. Large values of this measure occur when high eigenvalue ratios are observed for long time intervals, indicating that the timeseries is non-stochastic (denoting structure in the timeseries).
\end{itemize}

		\begin{figure}[ht]
			\centering
			\includegraphics[width=0.9\linewidth]{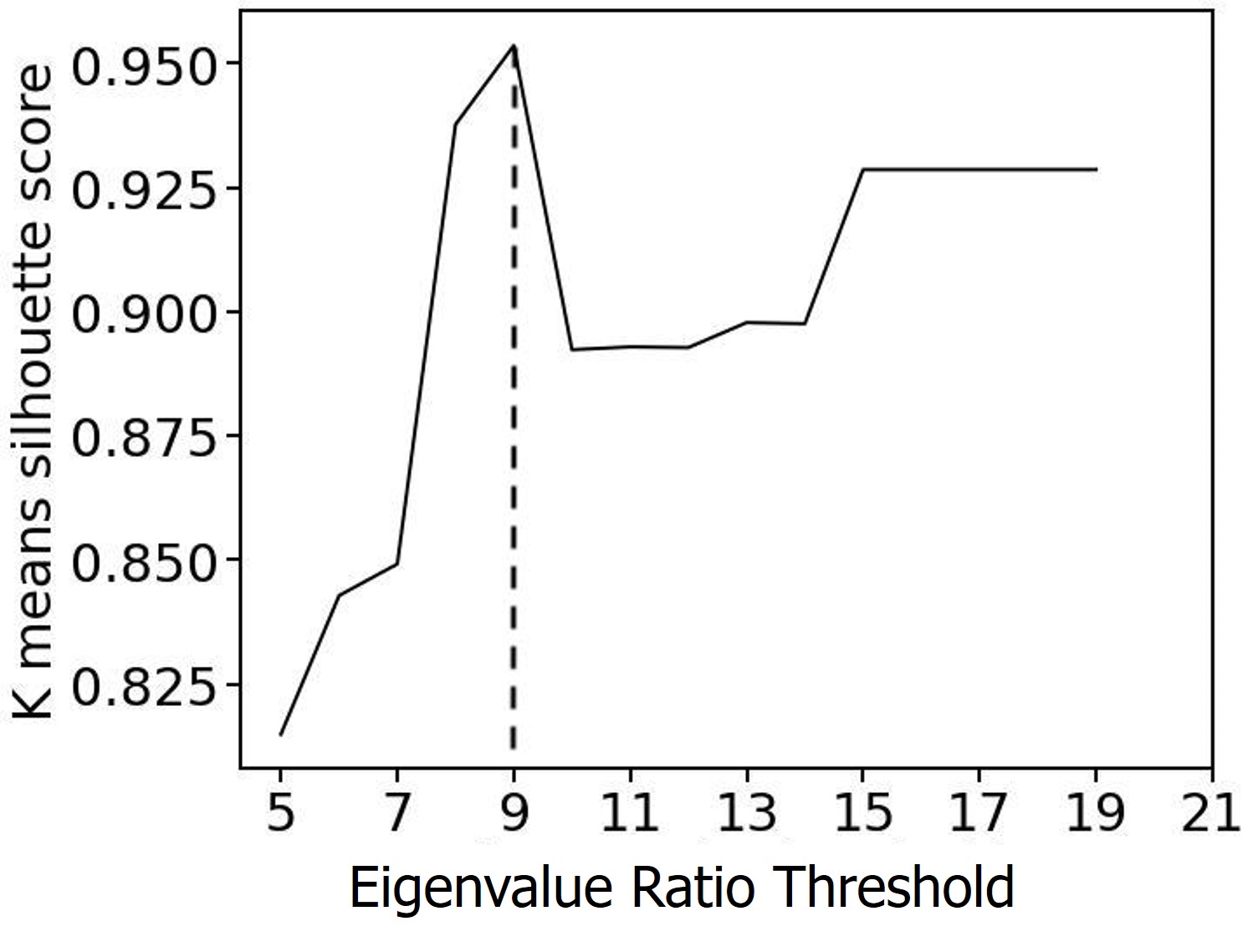}
			\caption{Silhoutte score vs. Eigenvalue ratio threshold. Maximum silhoutte score indicates the best clustering, which is obtained at a threshold of 9.}
			\label{sill}
		\end{figure}

In computing the above-described features, the value of the chosen threshold, ``Th" is critical. It is carefully chosen using the criterion described subsequently.

\textbf{Computing optimal value of threshold $Th$}: For optimal value of $Th$, we examine plot of the silhouette score of K-Means clustering, with $K=2$ (stochastic and non-stochastic clusters), performed using the computed features, as a function of the eigenvalue ratio threshold. The threshold that results in the best silhouette score is taken as $Th$, as shown in Fig. \ref{sill}. \\

\subsection{Computations Required} 
The required computation is the sum total of those incurred separately for the SVD decomposition based leg and with PCA decomposition-classification.
 
\textit{SVD decomposition and Topological descriptors:} The order of computations is determined by complexity of SVD-decomposition. It is well known that for a $k \times M$ matrix, the order of computations required is $O(k^2M)$, which is linear in $M$. Here the data matrix is of size $k\times M$, where $M \ll N$ ($N$ is the length of the timeseries). The complexity of computing topological descriptors is constant, since the binary image containing the E1 vs. E2 plots is of fixed size $k\times k$ (independent of the length of the timeseries $N$). Hence computational cost incurred is bounded by $O(N)$.

\textit{PCA-derived features and classification:} For a time series of length $N$, the computations are as follows: \begin{itemize}
	\item Covariance matrix [which involves computing $X^TX$,  where $X$ is of dimensions $ {\floor*{\frac{N}{2}}} \times 2$: $O(N)$].
	\item Computing eigenvalues of $2\times 2$ covariance matrix takes constant time.
	\item These computations are repeated for every iteration, where the maximum number of iterations possible is ${\rm log}N$ (assuming maximum number of splits).
	\item SVM binary classification takes constant time.
\end{itemize} 
 Hence the time complexity using PCA approach is $O(N{\rm log}N)$. 
 
Combining them both, the number of computations required is $O(N)+O(N{\rm log}N)$.

\section{Results} 
\subsection{On Synthetic Data}
Multiple realizations of the standard stochastic timeseries pertaining to white noise and pink noise are generated. For non-stochastic label, the timeseries from Lorentz system and logistic map (growth rate = 4) are considered. In all, 41 timeseries to represent the two labels in a fair manner are taken. 

\subsubsection{Interpreting Topological Descriptors} 
\begin{figure}
	\centering
	\includegraphics[width=\linewidth]{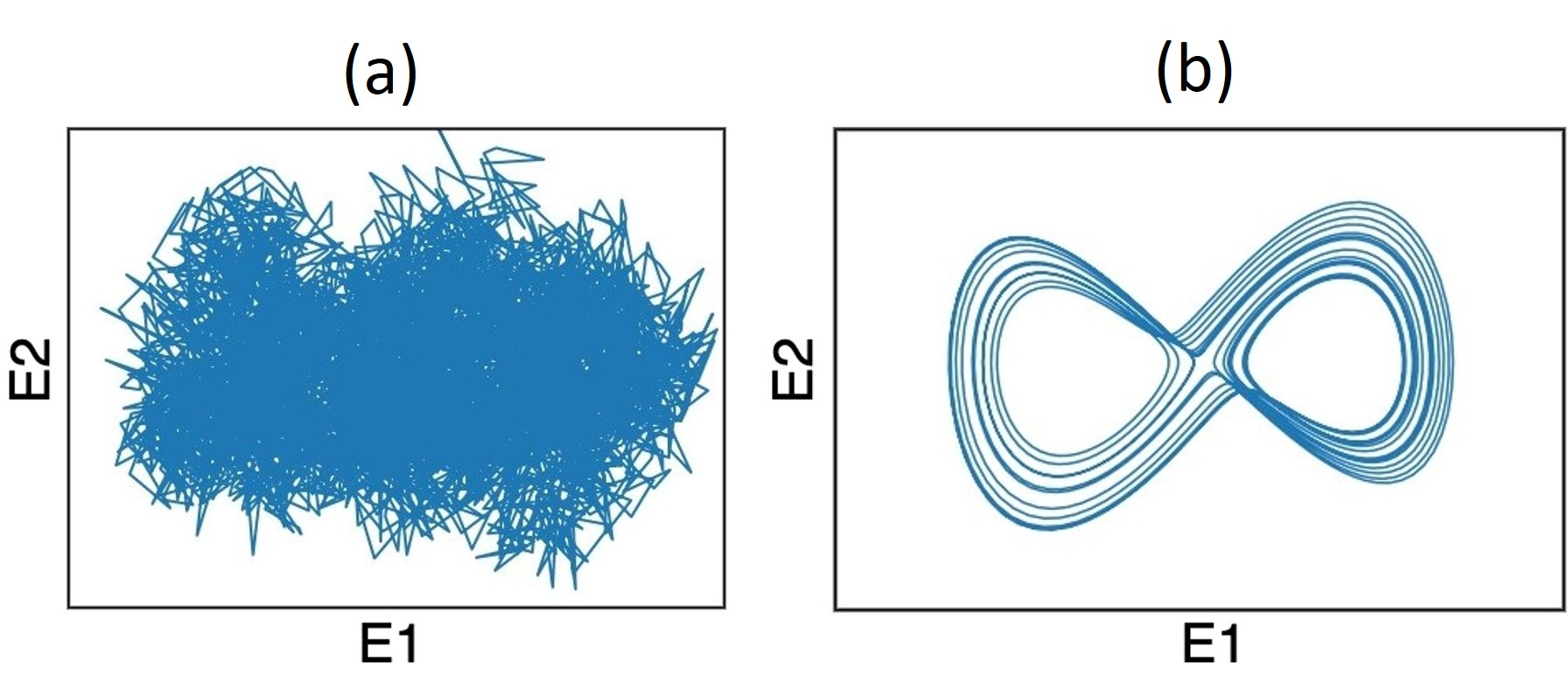}
	\caption{Comparison of E1 vs. E2 plots for synthetic data: (a) white noise (stochastic), (b) Lorentz system (non-stochastic).}
	\label{E1vsE2_wn_Lor}
\end{figure}

Results obtained using the SVD approach show, as expected, that for stochastic timeseries, the norm of the Betti descriptor is 1, while for non-stochastic timeseries it is always greater than 1. Thus the simple comparison on the norm of the Betti descriptor results in perfect classification. Representative plots of E1-E2 maps for stochastic label (white noise) has been provided in Fig. \ref{E1vsE2_wn_Lor}(a). It is evident from the plot that there is only one connected component without any void. Representative plots of E1-E2 maps for non-stochastic label (Lorentz system) has been provided in Fig. \ref{E1vsE2_wn_Lor}(b). Here, it is clearly observed that there are well defined voids. There is preferential directionality in this plot.

\begin{figure}
	\centering
	\includegraphics[width=\linewidth]{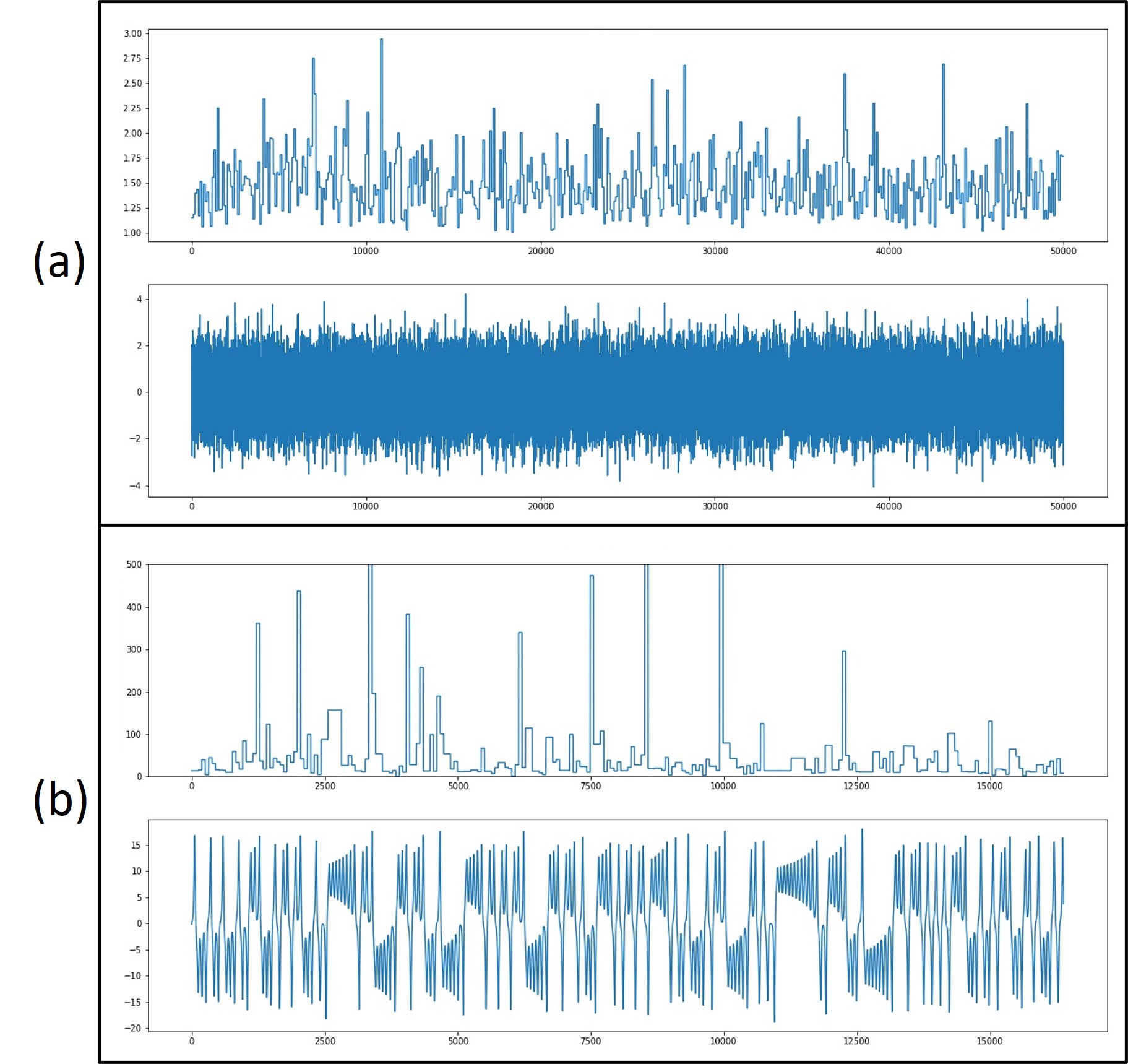}
	\caption{Comparison of eigenvalue ratios for synthetic data: (a) white noise (stochastic), (b) Lorentz system (non-stochastic). In each of the figures top panel consists of eigenvalue ratio plot while the bottom panel represents the timeseries.}
	\label{eigenvalue_ratio_wn_lorentz}
\end{figure}

%

%

\subsubsection{Interpreting PCA-derived Features} 

A comparison of eigenvalue ratios for representative stochastic (white noise) and non-stochastic (Lorentz system) labels, shown in Figs. \ref{eigenvalue_ratio_wn_lorentz}(a) and \ref{eigenvalue_ratio_wn_lorentz}(b) respectively illustrates the following: (i) The range of eigenvalue ratios for stochastic label is from 1 to 6 while that for non-stochastic label is 1 to 600. (ii) For a stochastic timeseries the eigenvalue ratio shows relatively more frequent fluctuations. In Figs. \ref{eigenvalue_ratio_wn_lorentz}(a) and \ref{eigenvalue_ratio_wn_lorentz}(b), top panel consists of eigenvalue ratio plot while the bottom panel represents the timeseries.

\begin{figure}
	\centering
	
		\centering
		\includegraphics[width=0.9\linewidth]{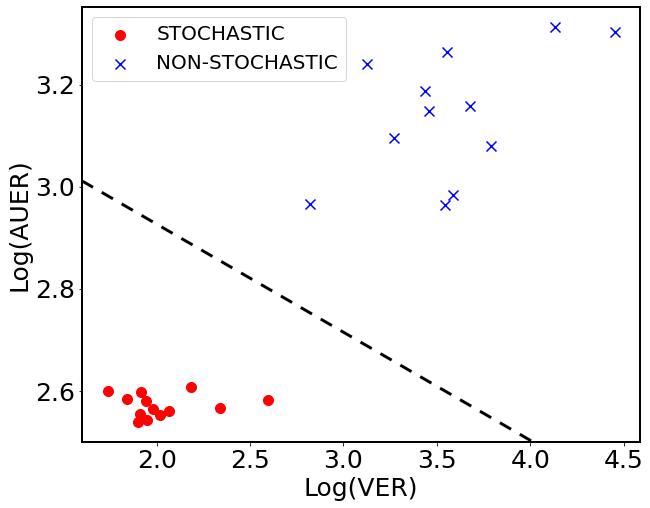}
		\caption{PCA-derived features mapping the timeseries to a linearly separable feature space - Illustrated on representative synthetic data. Log scale is used for better visualization. Dashed line indicates the classification boundary obtained using SVM classifier.}
		\label{svm_bound}
\end{figure}

The features (i) VER and (ii) AUER,  are computed for the considered synthetic signals. The scatter plot of these features is shown in Fig. \ref{svm_bound}. Since the dynamic range of values for the features VER and AUER is large, we use log-scale for the scatter plot. We observe the following: \begin{enumerate}
	\item VER: For a stochastic timeseries, since the variation in eigenvalue ratios is typically small, VER is also small. For a non-stochastic signal, since the eigenvalue ratios are diverse, VER is typically high.
	\item AUER: For a stochastic timeseries, since the eigenvalue ratios  are small across the entire length of the timeseries, AUER is also small. However, for a non-stochastic signal, the eigenvalue ratios remain high for longer time intervals. Hence AUER is significantly higher. 
\end{enumerate} 

It is observed that the features VER and AUER result in very small values for stochastic timeseries, while those for non-stochastic timeseries are relatively higher as expected. 

Figure \ref{svm_bound} shows that in this feature space, the two possible labels of timeseries, stochastic and non-stochastic, are linearly separable. Hence a linear SVM classifier is utilized. For training, computed features from 27 realizations of white noise and logistic map (growth rate = 4) are utilized. For validating the trained SVM, 14 realizations of pink noise and Lorentz system are used. The classification on all realizations of pink noise yields the label stochastic, while classification label for Lorentz system is obtained as non-stochastic, leading to perfect validation accuracy. This trained SVM is later used to classify real data.

Since concurrence in labels is obtained from both the legs of the analysis, the timeseries are labelled correspondingly. 

\subsection{On Real Data}
The proposed algorithm is applied on publicly available real data of a black hole source \textit{GRS 1915 + 105} \cite{rxte}. 12 temporal classes of timeseries, re-sampled with sampling interval of 0.1 second, are utilized. The length of timeseries, considered across 12 temporal classes, varies from a minimum of 16000 to a maximum of 34000. Data matrix $D$, as in equation (\ref{eqn:dmatrix}) is constructed with $K=5000$. From the SVD decomposition of $D$, we plot the top two right singular vectors (E1 vs. E2) for each of the 12 timeseries. Representative E1 vs. E2 plots for both the labels are shown in Fig. \ref{E1vsE2_gamma_kappa}. The observations made on synthetic data are seen to hold good even on real data. The obtained values of L1-norm of the Betti descriptor $\|B\|_1$, and the SVD classification labels, have been given in Table \ref{tab:results_1}. In order to compute PCA-derived features, VER and AUER, we first plot the eigenvalue ratio for the timeseries as illustrated in Figs. \ref{eigenvalue_ratio_kai_theta}(a) and \ref{eigenvalue_ratio_kai_theta}(b). The interval occupied by the eigenvalue ratio is observed to be similar to that observed over synthetic data for both the labels. It must be noted that the actual values of these intervals may not be consistent, however the order of these intervals are consistent. For stochastic timeseries, the intervals are in the range of 10s. For non-stochastic, the intervals range from 100s to 1000s. PCA-derived features, VER and AUER, are computed for each of these 12 timeseries as tabulated in Table \ref{tab:results_1}. These features are input to the SVM classifier to obtain the PCA classification label as seen in Table \ref{tab:results_1}. The coloumn ``Match" indicates the concurrence or lack of, between the two labels. The final coloumn declares the label of the corresponding timeseries. For all of the timeseries, except $\delta$, concurrence is obtained. The label for $\delta$ class is ``Uncertain" due to conflicting inferences, implying need for further investigation.

\section{Discussion} 
The proposed method utilizes matrix decomposition based techniques in two complementary ways to understand the given timeseries. Each of these analysis legs computes the label independently. 

\begin{table*}[t]
	\caption{Results on real data: The $\delta$ time series class for which SVD label and PCA label do not concur is labeled as ``Uncertain" (U)  ($NS$: Non-Stochastic and $S$: Stochastic)}
	\begin{center}
			\begin{tabular}{|C{0.8cm}|C{1.5cm}|C{1.5cm}|C{0.75cm}|C{0.9cm}|C{0.75cm}|C{0.9cm}|C{2cm}|}
			\hline
			Class  & Betti Norm & SVD Label & VER & AUER & PCA Label  &Match &Proposed Method Label\\
			\hline
			$\chi$ & 1 & S & 0.25 & 6.05 & S &Yes & S \\
			\hline
			$\gamma$ & 1 & S & 1 & 16 & S  &Yes & S\\
			\hline
			$\phi$ & 1 & S & 0.5 & 15 & S &Yes & S\\
			\hline
			$\delta$ & 1 & S & 9.7 & 26.2 & NS  & No & U \\
			\hline
			$\mu$  & 2 & NS & 51 & 12 & NS  & Yes & NS\\
			\hline
			$\nu$ & 7 & NS & 32 & 16 & NS & Yes & NS\\
			\hline
			$\alpha$  & 6 & NS & 1.9 & 27.7 & NS &Yes & NS\\
			\hline
			$\theta$ & 5 & NS & 778 & 58 & NS & Yes & NS\\
			\hline
			$\rho$ & 2 & NS & 147 & 35 & NS  & Yes& NS \\
			\hline
			$\beta$ & 4 & NS & 483 & 43 & NS & Yes & NS\\
			\hline
			$\kappa$  & 4 & NS & 5199 & 144 & NS &Yes& NS \\
			\hline
			$\lambda$ & 4 & NS & 6782 & 314 & NS & Yes & NS\\
			\hline

		\end{tabular}
		\label{tab:results_1}
	\end{center}
\end{table*}

\subsection{SVD Decomposition - Topological Analysis} 
\begin{figure}
	\centering
	\includegraphics[width=\linewidth]{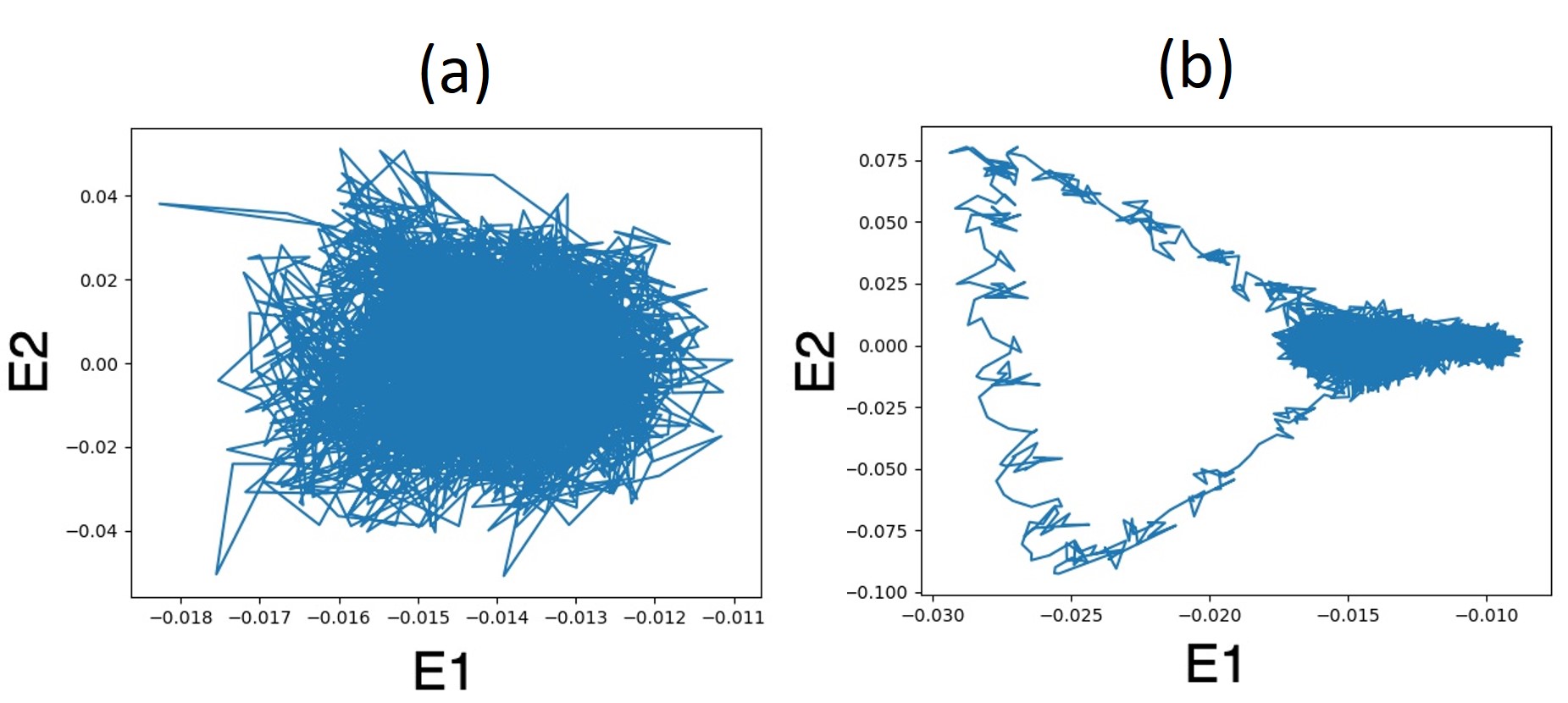}
	\caption{Comparison of E1 vs. E2 plots for real data: (a) $\gamma$ class  (stochastic), (b) $\kappa$ class (non-stochastic), of \textit{GRS 1915 + 105}.}
	\label{E1vsE2_gamma_kappa}
\end{figure}

SVD decomposition of the constructed data matrix as described in equation (\ref{eqn:dmatrix}), yields singular vectors that contain information of the temporal dynamics. The plot of the top two singular vectors (E1 vs. E2 map) is observed. As hypothesized, in case of a stochastic timeseries, as in Fig. \ref{E1vsE2_gamma_kappa}(a), since there is no preference in orientation, the tuples in the E1 vs. E2 plot resemble a solid blob. As expected, the Betti description of this plot would result in $B_{0}=1, B_{1}=0$, since there would be only one connected component, with no voids. This is because there is no preferred orientation due to lack of structure in the stochastic timeseries. Hence the norm of the Betti descriptor, $\left \| B \right \|_1=1$, for a stochastic timeseries. In contrast, for a non-stochastic timeseries, as in Fig. \ref{E1vsE2_gamma_kappa}(b), the plot of the top two singular vectors (E1 vs. E2 plot) that captures temporal dynamics, demonstrates a definite preference for certain orientation(s) and results in one or multiple connected components with well-defined voids. This leads to a value of  $B_{0} \ge 1$, since there could be multiple connected components. However, $B_{1} > 1$, since the structure in the timeseries would guide the trajectory of the singular vectors preferentially, leaving behind voids. Hence $\left \| B \right \|_1 > 1$ for a non-stochastic timeseries. 

%
\subsection{PCA-derived Features - SVM Classification}

\begin{figure}
	\centering
	\includegraphics[width=\linewidth]{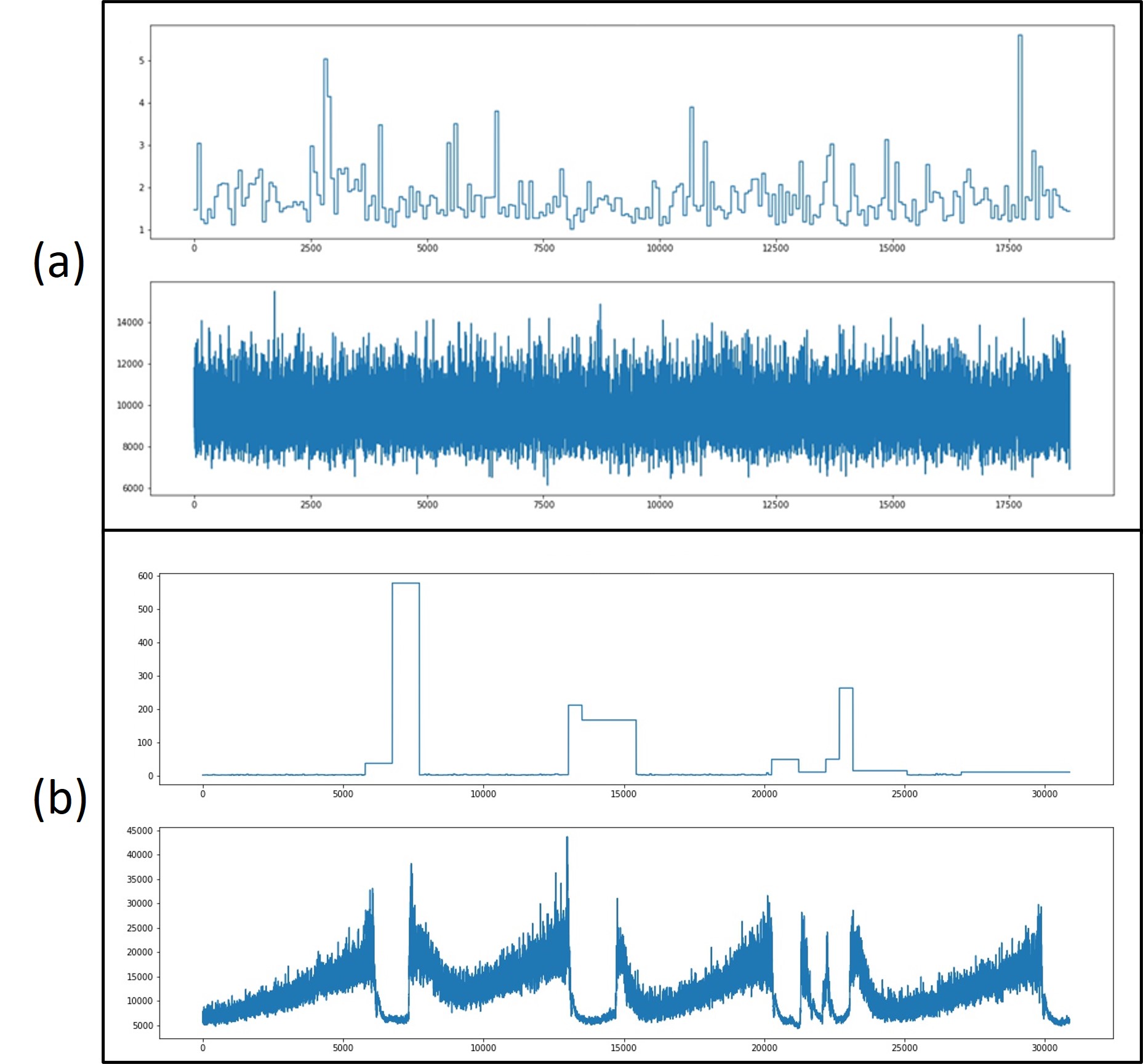}
	\caption{Comparison of eigenvalue ratios for real \textit{GRS 1915 + 105} data: (a) $\chi$ class (stochastic), (b) $\theta$ class  (non-stochastic). In each of the figures top panel consists of eigenvalue ratio while the bottom panel represents the timeseries.}
	\label{eigenvalue_ratio_kai_theta}
\end{figure}

PCA decomposition does not utilize temporal ordering which is complementary to SVD decomposition which utilizes temporal ordering. In PCA decomposition, the temporal order is overlooked when computing moments (mean, variance, co-variance) of the data segments split across successive iterations. For a stochastic timeseries, the ratio of eigenvalues varies within a small range of values, since there is no preference in direction. However, for a non-stochastic timeseries, the ratio of eigenvalues varies significantly, since there is a preferred orientation owing to the structure. This comparison is seen in Figs. \ref{eigenvalue_ratio_kai_theta}(a) and \ref{eigenvalue_ratio_kai_theta}(b). This observation is exploited in constructing the PCA-derived features of VER and AUER. The SVM trained on synthetic data has been utilized to classify the real data. The perpendicular distance of the feature vector from the separating plane can be interpreted as the confidence in the classification label.

%

 The SVD-based approach relies on the physical phenomenon in the construction of the data matrix. In contrast, PCA-based analysis has no requirement for any such phenomenon-specific inputs. Also, SVD regards temporal ordering of data in the construction of data matrix, while PCA treats the data temporally-agnostic within each temporal segment, while computing moments such as mean, variance and covariance. The two complementary approaches, result in analyzing the timeseries with different perspectives. Hence when the obtained labels on both the legs concur, the timeseries is declared as such. Else, the timeseries is declared ``Uncertain".

Here, we also compare results obtained using the proposed method, with those in literature. The obtained results are compared with the most recent deep learning framework based approaches, called the DS-measure \cite{ds_icassp}, the LSS technique \cite{lss} and the traditional CI approach \cite{Adegoke2018}. Each of the methods uses a different perspective to understand the timeseries. Hence it is important to incorporate findings from various perspectives before conclusively arriving at an inference. 

\subsection{Comparison With Deviation from Stochasticity (DS) Approach}
DS approach \cite{ds_icassp} utilizes deep learning framework, performing multi-scale analysis, to determine if the given timeseries is stochastic or non-stochastic. The approach computes a measure called ``DS", which has been illustrated to be smaller for stochastic timeseries and relatively larger for non-stochastic timeseries. DS measure consists of two intermediate factors $CV_1$ and $CV_2$, where $CV_1$ measures the impact of multiple timescales and $CV_2$ measures temporal variation within the timescales. In the Table \ref{DS_compare}, the classification obtained on the 12 temporal timeseries classes of black hole \textit{GRS 1915+105}, are compared with those obtained using the proposed methodology. It is observed that results obtained using the proposed approach are the same as that obtained using DS approach on all the timeseries, except for one. The $\delta$ timeseries is classified here as ``Non-stochastic" as reported in \cite{ds_icassp}. 

\begin{table}[h]
	\centering
	\small
	\caption{Comparison with DS measure on black hole source \textit{GRS 1915 + 105} data (S:Stochastic, NS:Non-stochastic, U:Uncertain)}
	\centering
	\label{DS_compare}
	\begin{tabular}{| C{1.6cm}| C{0.8cm} | C{0.8cm} | C{2cm} | }
		\hline
		\textbf{Timeseries}  &\textbf{$DS$} &DS Label &Proposed Method Label \\ \hline
		$\chi$			&0.18 &S &S \\ \hline
		$\gamma$				&0.83 &S &S \\ \hline
		$\phi$			&0.97 &S &S \\ \hline
		\textbf{$\delta$}	&\textbf{4.07} &\textbf{NS} &\textbf{U} \\ \hline
		$\mu$				&4.67 &NS &NS \\ \hline
		$\nu$				&4.79 &NS &NS \\ \hline
		$\alpha$			&5.16 &NS &NS \\ \hline
		$\theta$			&6.41 &NS &NS \\ \hline
		$\rho$				&7.90 &NS &NS \\ \hline
		$\beta$				&8.82 &NS &NS \\ \hline
		$\kappa$			&11.96 &NS &NS \\ \hline
		$\lambda$			&12.83 &NS &NS \\ \hline
	\end{tabular}                                              
\end{table}

\subsubsection{Interpreting DS Measures} It is observed that the DS measures obtained for the temporal classes with ``Stochastic" label are relatively small, below the threshold, as determined in \cite{ds_icassp}. Out of the temporal classes, declared as ``Non-stochastic", the temporal class, $\delta$, has the lowest DS measure. 

On breaking down the DS measure obtained for $\delta$ timeseries into its contributing factors, $CV_1$ and $CV_2$, it is observed that $CV_1$ has a high value, while $CV_2$ has a low value, as the evidence of contradiction.  This points to the inference that $\delta$ timeseries is a modulation of stochastic and non-stochastic timeseries. This observation is in line with the description of the $\delta$ timeseries in \cite{Belloni}, that this specific timeseries is described to be similar to Red-noise (PSD decays quadratically with frequency), and also exhibits structured dips that occur all through the timespan. These two contradicting aspects are captured in the high value of $CV_1$ and low value of $CV_2$, as the evidence of modulation of stochastic and non-stochastic factors. In this case,  DS measure (product of $CV_1$ and $CV_2$) turns out to be high, since $CV_1$ $>>$ $CV_2$. Hence this is labeled as non-stochastic by the DS method.

\subsection{Comparison with Latent Space Signature (LSS) approach}
LSS approach \cite{lss} utilizes deep learning frame-work to convert a 1D timeseries to a 2D representation. Towards this, autoencoders, that analyse the given timeseries in both time- and frequency-domains, are utilized. The LSS representations are utilized to create the corresponding binary LSS images. These LSS images are fed to a classifier to obtain the classification label corresponding to the timeseries. 

In the Table \ref{LSS_compare}, the classification obtained on the 12 temporal timeseries classes of black hole \textit{GRS 1915+105}, are compared with those obtained using the proposed methodology. It is observed that results obtained using proposed approach is the same as that obtained using LSS approach on 11 timeseries, except for $\delta$ timeseries.

\begin{table}[h!]
	\centering
	\small
	\caption{Comparison with LSS label on black hole source \textit{GRS 1915 + 105} data (S:Stochastic, NS:Non-stochastic, U:Uncertain)}
	\centering
	\label{LSS_compare}
	\begin{tabular}{| C{1.6cm}| C{1cm} | C{1cm} | C{0.8cm} | C{2cm} | }
		\hline
		\textbf{Timeseries} &\textbf{$C_S$} &\textbf{$C_{NS}$} &LSS Label &Proposed Method Label\\ \hline
		$\chi$			&0.9898	&0.0102 &S &S \\ \hline
		$\gamma$		&1.00	&0.00 &S &S \\ \hline
		$\phi$			&1.00	&0.00 &S &S \\ \hline
		$\delta$&0.807	&0.193 &S&U \\ \hline
		$\mu$			&0.0041	&0.9959 &NS &NS \\ \hline
		$\nu$			&0.0779	&0.9221 &NS &NS \\ \hline
		$\alpha$		&0.0055 &0.9945 &NS &NS \\ \hline
		$\theta$		&0.1206	&0.8794 &NS &NS \\ \hline
		$\rho$			&0.1593	&0.8407 &NS &NS \\ \hline
		$\beta$			&0.0083	&0.9917 &NS &NS \\ \hline
		$\kappa$		&0.0102	&0.9898 &NS &NS \\ \hline
		$\lambda$		&0.0043	&0.9957
		&NS &NS \\ \hline
	\end{tabular}                                              
\end{table}  

\subsubsection{Interpreting LSS Confidence Scores} The confidence scores, $C_{S}$ and $C_{NS}$  for ``Stochastic" and ``Non-stochastic", respectively, tabulated in Table \ref{LSS_compare}, measure the strength of the corresponding label. For temporal classes such as $\chi$, $\gamma$, $\phi$, the confidence score for the label ``Stochastic" is 1 or very close to 1. However, for the temporal class $\delta$, the confidence score for label ``Stochastic" is relatively less, at 0.8. This points to decreased confidence in classifying this particular timeseries. On the other hand the temporal classes that have been declared as ``Non-stochastic", have confidence scores higher than 0.84. It is also observed that lower ``Non-stochastic" confidence scores are obtained for $\rho$ and $\theta$. From the table it is evident that the temporal class whose confidence score is least is the $\delta$ timeseries. 

As evident from results obtained using the proposed method, the ambiguity regarding the $\delta$ timeseries persists. Besides, the results obtained using the DS-measure also point to the confusion in the label of this specific timeseries. It is worth noting that two different approaches that utilize deep learning framework result in contradictory labels for the $\delta$ timeseries. Hence, the proposed approach labels as ``Uncertain".

\subsection{Comparison with Correlation Integral (CI) Approach}
CI \cite{Adegoke2018} analyses the timeseries by determining the CD to understand the underlying physical process. CI works by clustering of points with an appropriate choice of neighborhood. This approach requires several iterations and possible computation of additional parameters such as Lyapunov exponent in order establish a timeseries as stochastic or non-stochastic. However, the proposed methodology can identify the label in one single run, and also provide the confidence in the classification label. In Table \ref{CI_compare}, the classification obtained on the 12 temporal timeseries classes of black hole \textit{GRS 1915+105}, are compared with those obtained using the proposed methodology. It is observed that results obtained using the proposed approach is the same as that obtained using CI approach on all the timeseries except for $\delta$ timeseries. 

\begin{table}
	\centering
	\small
	\caption{Comparison with CI label on black hole source \textit{GRS 1915 + 105} data (S:Stochastic, NS:Non-stochastic, U:Uncertain)}
	\centering
	\label{CI_compare}
	\begin{tabular}{| C{1.6cm}| C{0.8cm} | C{2cm} |}
		\hline
		\textbf{Timeseries}  &CI Label &Proposed Method Label\\ \hline
		$\chi$			&S &S \\ \hline
		$\gamma$				&S &S \\ \hline
		$\phi$			&S &S \\ \hline
		\textbf{$\delta$} &\textbf{NS} &\textbf{U} \\ \hline
		$\mu$				 &NS &NS \\ \hline
		$\nu$				 &NS &NS \\ \hline
		$\alpha$			 &NS &NS \\ \hline
		$\theta$			&NS &NS \\ \hline
		$\rho$				&NS &NS \\ \hline
		$\beta$				&NS &NS \\ \hline
		$\kappa$			&NS &NS \\ \hline
		$\lambda$			&NS &NS \\ \hline
	\end{tabular}                                              
\end{table}  

\subsubsection{Interpreting CI Labels} CI approach estimates CD, towards determining the label of the given  timeseries. If CD saturates, the timeseries is considered to be non-stochastic, else it is considered to be stochastic. As per authors in \cite{Adegoke2018}, the temporal classes with low saturated values of CD were labeled Non-stochastic, while the others were labeled stochastic. Obtaining these labels requires repeated experiments with varying values of Embedding Dimension. In order to label a timeseries as stochastic, where the value of CD turns out to be approximately equal to Embedding Dimension, several more experiments are required to illustrate that there is no saturation in CD. However, a big advantage of this approach is that it gives insight into the complexity of the dynamics through the value of the CD. 

\section{Conclusion}

Exploring different techniques, in order to have a conclusive inference for black hole systems, turns out to be indispensable. We propose a decomposition-based algorithm, that utilizes two separate legs (SVD- and PCA-based) each of which determines an independent label for a timeseries, using complementary approaches. If the labels obtained using the two legs happen to be the same, the timeseries is declared as such; but in case there is a contradiction, the label is deemed to be ``Uncertain". The performance of the proposed approach is illustrated on 12 temporal classes of timeseries of black hole \textit{GRS 1915+105} obtained from \textit{RXTE} satellite data. We compare inferences of the proposed method with two available deep-learning based frameworks as well as the traditional CI-based approach. Of the 12 temporal classes,  concurrence across all approaches is obtained on 11 of them. However, on one temporal class, it yields the label ``Uncertain". The same timeseries is observed to have obtained conflicting labels using the deep-learning approaches as well. This points to needing further investigation towards understanding the black hole dynamics.


\bibliographystyle{IEEEbib}
\bibliography{refs}



\end{document}